\def\year{2022}\relax
\documentclass[letterpaper]{article} 
\usepackage{aaai22}  
\usepackage{times}  
\usepackage{helvet}  
\usepackage{courier}  
\usepackage[hyphens]{url}  
\usepackage{graphicx} 
\usepackage{amssymb}
\usepackage{natbib}  
\usepackage{caption} 
\DeclareCaptionStyle{ruled}{labelfont=normalfont,labelsep=colon,strut=off} 
\usepackage{booktabs}
\usepackage[load-configurations=version-1]{siunitx} 
\usepackage{color,soul}
\usepackage{graphicx}
\usepackage{multirow}
\usepackage{subfigure}
\usepackage{balance}
\usepackage{threeparttable}
\usepackage{multirow}
\usepackage{xcolor}
\usepackage{color}
\usepackage{soul}
\usepackage{etoolbox}

\newcommand{\acc}{\textit{Accuracy }} 
\newcommand{\Uncertainty}{\textit{Uncertainty }} 
\newcommand{\ECE}{\textit{ECE }} 
\newcommand{\Brier}{\textit{Brier }} 

\definecolor{magicmint}{rgb}{0.67, 0.94, 0.82}

\title{Benchmarking Uncertainty Quantification \\ on Biosignal Classification Tasks under Dataset Shift}

\author{%
 Tong Xia,
  Jing Han,
  Cecilia Mascolo\\
  }
\affiliations{Department of Computer Science and Technology, University of Cambridge, UK\\
 tx229@cam.ac.uk}

\begin{document}

\maketitle

\begin{abstract}


A biosignal is a signal that can be continuously measured from human bodies, such as respiratory sounds, heart activity (ECG), brain waves (EEG), etc, based on which, machine learning models have been developed with very promising performance for automatic disease detection and health status monitoring. However, dataset shift, i.e., data distribution of inference varies from the distribution of the training, is not uncommon for real biosignal-based applications. To improve the robustness, probabilistic models with uncertainty quantification are adapted to capture how reliable a prediction is. Yet, assessing the quality of the estimated uncertainty remains a challenge. In this work, we propose a framework to evaluate the capability of the estimated uncertainty in capturing different types of biosignal dataset shifts with various degrees. In particular, we use three classification tasks based on respiratory sounds and electrocardiography signals to benchmark five representative uncertainty quantification methods. Extensive experiments show that, although Ensemble and Bayesian models could provide relatively better uncertainty estimations under dataset shifts, all tested models fail to meet the promise in trustworthy prediction and model calibration. Our work paves the way for a comprehensive evaluation for any newly developed biosignal classifiers.



\end{abstract}

\section{Introduction}
\label{sec:intro}
The growth of commercial wearables and the ubiquity of smartphones with numerous sensors have enabled multi-modal, affordable, non-invasive, round-the-clock biosignal collection~\citep{athavale2017biosignal}, based on which deep learning facilitates a wide spectrum of mental and physical health applications. 
Despite the impressive performance achieved by deep learning, the underlying premise is that training and test data are from the same distribution. Yet, this assumption often does not hold in practice, because dataset shift caused by user variability, device discrepancy, artefact, and other factors are ineluctable during in-the-wild biosignal acquisition~\citep{pooch2019can}.
Most existing deep learning models cannot flag this distributional shift and tend to be over-confident during inference, which may, in the long run, undermine people's trust in applying deep learning for healthcare~\citep{guo2017calibration}. 



In the field of machine learning,  predictive uncertainty has been used as a measurement of how a deep neural network can be trusted~\citep{gawlikowski2021survey}. Different types and sources of uncertainties have been identified and a variety of approaches to quantify uncertainty in neural networks have been proposed~\citep{abdar2021review}. Given the importance of risk management for safety-critical health applications, uncertainty is recognised to be helpful, because it can not only inform the confidence of the prediction but also provide the opportunity to keep doctors in the loop to correct the potentially wrong automatic predictions. \citeauthor{leibig2017leveraging} found that in the task of diagnosing diabetic retinopathy from fundus images of the eye, incorrect prediction usually yielded higher uncertainty than true predictions: by excluding the least uncertain predictions, the automatic diagnosis' accuracy could be improved from 87\% to 96\%~\citep{leibig2017leveraging, singh2021uncertainty}.

In spite of the efforts to estimate and understand the uncertainty of health-related deep learning models, key aspects are still under-explored. \textit{First}, most of the previous work focuses on clinical images~\citep{gawlikowski2021survey}, while how uncertainty performs on other health physical signals, e.\,g., biosignal from wearables, remains unclear.
\textit{Moreover}, the quality of uncertainty is mainly assessed in the independent and identically distributed (i.i.d.) testing set, while in the distributional shifted regime, problems like how trustworthy the existing uncertainty estimation methods are unsolved.
 
To answer these questions, in this paper, we conduct a comprehensive evaluation across five uncertainty quantification methods on three representative biosignal classification tasks under a controlled dataset shift. 
The implemented uncertainty quantification methods cover Bayesian neural network, approximate Bayesian neural network, calibration and deep ensemble approaches, and the tasks include audio-based COVID-19 prediction, breathing-based respiratory abnormality prediction, and ECG-based heart arrhythmia detection applications. 
To assess the quantified uncertainty, we propose a framework for analysing the above methods on dataset shift without requiring the collection of new datasets. Specifically, the key mechanism is to empirically synthesise signal-specific distributional shift according to real signal data collection scenarios, so that both the shift type and the degree can be controlled and the evaluation framework can be generalised to any biosignal tasks. Consequently, we find that the widely used uncertainty estimation approaches fail to yield well-calibrated uncertainty under dataset shift, and we draw attention to better methods for safety-critical health applications.

\section{Related Work}
\label{sec:related}
Uncertainty is recognised as a measurement of model's trust and its importance has been widely discussed in the literature, particularly for deep learning-enabled health applications~\citep{bhatt2021uncertainty,moon2020confidence}. The quantified uncertainty can be used for selective prediction: keeping low-uncertain outputs but referring high uncertain (unsafe) predictions to doctors, which allows clinicians in the loop and improves the system robustness. Diagnosing diabetic retinopathy from fundus images of the eye is a commonly used task to assess the estimated uncertainty~\citep{leibig2017leveraging,singh2021uncertainty,van2020uncertainty,raghu2019direct,wang2021beyond}, and uncertainty-aware lung disease detection from X-rays also gains massive attention~\citep{ghoshal2020estimating,singh2021uncertainty}. To the best of our knowledge, only a few works explored uncertainty for biosignal~\citep{singh2021uncertainty, xia2021uncertainty}.
Moreover, all those works demonstrate the quality of uncertainty in the testing set which was selected from the same distribution with training data,  while the effectiveness remains unclear under dataset shift.

As dataset shift is now being highlighted frequently by the research community, there have been some attempts \cite{ovadia2019can,schwaiger2020uncertainty,ulmer2020trust,band2021benchmarking} to assess the estimated uncertainty (none of them is for biosignals): 
\citeauthor{ovadia2019can} conducted a comparison of multiple uncertainty estimation methods on image, context, and ad-click datasets, discovering that along with accuracy, the quality of uncertainty consistently degrades with increasing dataset shift regardless of method. \citeauthor{ulmer2020trust} proved uncertainty estimation does not enable reliable out-of-distribution detection on medical tabular data. 
Recently, \citeauthor{band2021benchmarking} proposed a benchmark to evaluate Bayesian deep learning on diabetic retinopathy detection task, where two human retina image datasets with country shift and severity shift are constructed for use. 
\textit{Inspired by but different from the aforementioned works, in this paper, we aim to close the gap between uncertainty quantification and biosignal dataset shift, towards more realistic performance evaluation and more reliable healthcare deployment.}
 
\section{Uncertainty Quantification}
\label{sec:method}


\subsection{Notation and Problem Setup}
Let $x \in \mathbb{R}^{d}$ represent a set of d-dimensional features and $y \in \{1, . . . ,k\}$ denote corresponding labels (targets) for $k$-class classification. We assume that a training dataset $\mathcal{D}$ consists of \textit{i.i.d.} samples $\mathcal{D} = \{(x_n, y_n)\}^N_{n=1}$.
 We focus on classification problems, so we use a neural network to model $p_{\theta}(y|x)$ and estimate the parameters $\theta$ using the training dataset. During inference, we evaluate the model predictions against a testing set $\mathcal{D'} = \{(x', y')\}$, sampled from a shifted distribution from $D$. 

\subsection{Uncertainty Estimation Approaches}
For model $\theta$, the output probability $p_{\theta}$ from $Softmax$ at hand may indicate the confidence of the prediction to some degree. However, it tends to overestimate the confidence and requires further calibration~\citep{guo2017calibration}. Only well-calibrated uncertainty would be very useful to tell to what degree the model is certain about its predictions. Two distinct types of uncertainties exist: \textit{aleatoric uncertainty} stems from stochastic variability inherent in the data generating process (also know as \textit{data uncertainty}), while \textit{epistemic uncertainty} arises due to our lack of knowledge about the data generating mechanism. More specifically, \textit{epistemic uncertainty} is associated with model structures\&parameters (also known as \textit{model uncertainty}), and the systematic discrepancy between the training and testing data (\textit{distributional shift})~\citep{liu2019accurate}. 
Existing uncertainty quantification methods mainly include the following categories, 
\vspace{4pt}
\\\textbf{Bayesian Methods.}
 Bayesian methods explicitly define a combination of infinite plausible models according to a learnable prior distribution estimated from the observed data. Given the observed data set $\mathcal{D} = \{(x, y)\}$, a conditional probability  $p(y|x,\theta)$ indicates how each model $\theta$ explains the relation of input $x$ and target $y$. With the posterior distribution $p(\theta|D)$, for a new test data point $x'$, the predictive posterior distribution can be derived as,
 \begin{equation}
     p(y|x') =  \int p(y|x',\theta) p(\theta|D) d\theta,
\end{equation}     
\begin{equation}
     p(\theta|D) = \frac{p(D|\theta)p(\theta)}{ \int p(D|\theta) p(\theta) d\theta},
 \end{equation}\label{equ:bayesian}
where $p(D|\theta)p(\theta)=\prod_np(y_n|x_n,\theta)$.
\vspace{2pt}
\\\textbf{Monte Carlo Dropout.} The above Bayesian methods, although can estimate uncertainty, will raise additional and expensive computations. 
In general, dropout is a widely used technique during training to tackle overfitting and it is switched off at inference time~\citep{baldi2013understanding,srivastava2014dropout}. However, \cite{gal2016dropout} leveraged variational distribution to view dropout in the forward phrase as approximate Bayesian inference. In this manner with dropout kept during inference, the predictive probability can be computed through the randomly sampled models and treated as an estimation of the uncertainty. More importantly, it involves no extra computational cost during training.    
\vspace{4pt}
\\\textbf{Ensemble Methods.}
Ensemble, also known as frequentist methods, does not specify a prior distribution over parameters. Instead of learning infinite models, ensemble approaches only require a limited number of models, which is computationally tractable~\citep{ganaie2021ensemble}. Herein, the final predictive probability for each instance $x'$ is estimated by simply averaging the outputs over all $M$ models:
\begin{equation}
     p(y|x') = \frac{1}{M} \sum_{i=1}^{M} p(y|x',\theta^i).
\end{equation}\label{equ:ensemblem}
\vspace{-5pt}
\\\textbf{Model Calibration.}
The probability that a model outputs should reflect the true correctness likelihood. However, most modern neural networks are poorly \textit{uncertainty calibrated}~\citep{kumar2019verified}. In this respect, post-hoc probability calibration is reported to be helpful to alleviate the gap between outputs and the true likelihood, although it cannot capture model uncertainty. From the literature~\citep{guo2017calibration}, the simplest and most straightforward approach for practical setting is temperature scaling: using one parameter to re-scale logits before passing them into softmax. 

To sum up, we select five methods for uncertainty estimation considering their prevalence, scalability, and practical applicability~\citep{ovadia2019can}.  They are, 
\begin{itemize}
    \item \textit{\textbf{Vanilla}}: Maximum softmax probability directly from the deterministic backbone model output.
    \item \textit{\textbf{Scaling}}: Post-hoc calibration from  vanilla probabilities by temperature scaling parameterised by value $T$~\citep{guo2017calibration}.
    \item \textit{\textbf{MCDropout}}: Monte-Carlo Dropout with a dropout rate of $p$ during inference~\citep{gal2016dropout}. A sample will be fed into the model $M$ times to quantify the model uncertainty.
    \item \textit{\textbf{Bayesian}}: Stochastic variational Bayesian inference~\citep{graves2011practical} with Gaussian priors.
    \item \textit{\textbf{Ensemble}}: Ensembles of several networks with identical structures, which are trained with random initialisation ~\citep{lakshminarayanan2017simple}. 
\end{itemize}

\section{Bencmark Tasks and Datsets}
\label{sec:exp}
 
\begin{table*}[h]
\centering
\resizebox{0.99\textwidth}{!}{
\begin{tabular}{llllc}
\toprule
\textbf{Task}          & \textbf{Dataset (Size)} & \textbf{Classes} & \textbf{Backbone} & \textbf{Accuracy}  \\ \cmidrule(r){1-5}
COVID-19      & Audio dataset (1,000 users)  & COVID-19 positive/negative   & VGGish      & 0.68\\
Respiratory   & Breathing recordings (1,990 clips) & Normal/abnormal     & ResNet      &0.75\\
Arrhythmia    & ECG (8,224 recordings)  & Normal/AF(atrial fibrillation)/others & Transformer &0.83 \\
\bottomrule
\end{tabular}}
\caption{A summary of the tasks, datasets, models, and baseline performance.}\label{Tab:tasks}
\end{table*}
 
\subsection{Biosignal classification tasks}
 From the literature, we select three representative health tasks with different biosignal modalities to investigate whether the estimated uncertainty works when dataset shift occurs.
\vspace{2pt}
\\\textbf{COVID-19 prediction.} Sound-based COVID-19 prediction has shown great promise in achieving affordable COVID-19 screening through smartphones~(\cite{imran2020ai4covid,brown2020exploring}). We choose to implement the model proposed in ~\citep{han2021sounds} for its available code and dataset\footnote{\url{https://www.covid-19-sounds.org/en/}}. This is a binary classification task, where cough, breathing, and voice sounds are transferred into spectrograms to distinguish COVID-19 positive from negative participants. VGGish~\citep{hershey2017cnn} is leveraged as the backbone model.
\vspace{2pt}
\\\textbf{Respiratory abnormality detection.} Auscultation of the lung is an important part of the clinical patient examination and is helpful in diagnosing various respiratory disorders. It is vital to distinguish abnormal respiratory sounds from normal ones to enable the correct treatment. The \textit{ICBHI 2017 Challenge}\footnote{\url{https://bhichallenge.med.auth.gr/}} provided a breathing sound database collected from heterogeneous auscultation equipment. A binary classification task is formulated to detect whether a breathing sound segment contains abnormalities, including crackle and wheeze. We use the backbone of a deep convolutional model (ResNet) proposed in~\cite{gairola2020respirenet} for its favourable performance.
\vspace{2pt}
\\\textbf{Heart arrhythmia detection.} ECG, another type of widely used biosignal, involves the recording of electrical impulses generated by the heart muscle during beating activity. Through ECG, arrhythmia (irregular beat) can be identified. For the evaluation, we use 
the dataset from~\citep{goldberger2000physiobank} and the developed transformer-based neural network in the \textit{PhysioNet Computing in Cardiology Challenge 2017}~(\cite{clifford2017af}) considering its high research impact\footnote{\url{https://physionet.org/content/challenge-2017/1.0.0/sources/#files-panel}}. The task is to predict, from a single short ECG lead recording (between 30\,s and 60\,s in length), whether the recording shows normal sinus rhythm, atrial fibrillation (AF), or an alternative rhythm. 

Table~\ref{Tab:tasks} presents a summary of the above tasks.  More details are introduced in Appendix~\ref{apd:task}.
 
\subsection{Dataset Shift}
To assess the quality of the predictive uncertainty yielded by different methods, we propose and apply the following perturbations covering major potential shifts in practice: 
\begin{itemize}
    \item \textbf{\textit{Mixing Gaussian noise.}}  Gaussian noise is statistical noise having a probability density function equal to the normal distribution. This type of noise is common in various types of signals and can arise from acquisition (e.\,g., sensor noise) or transmission (e.\,g., electronic circuit noise). We add the generated Gaussian noise to raw biosignals by controlling the SNR (signal-to-noise ratio, please refer to Appendix for details).
    \item \textbf{\textit{Mixing background noise.}} Random background or environmental noise is another type of prevalent noise when collecting biosignals, particularly in audio signals. To simulate this, we mix pre-recorded TV news with raw audio signals according to different SNRs (refer to Appendix).
    \item \textbf{\textit{Signal amplitude distortion.}}  The amplitude distortion, also called clipping, is the result of ``over-driving" the input of the amplifier~\citep{liang1999nonlinear}, which is a part of the analogy signal acquisition circuit. For example, when a user speaks closed to the microphone loudly, the audio signal can be distorted with some peak value flattened in the waveform view. We synthesise the distortion by replacing the amplitude over a pre-defined threshold to the threshold value.
    \item \textbf{\textit{Signal segment missing.}}
    When signal acquisition or transmission is unstable, some segments can be missing, which leads to incomplete signals in the time domain. According to this, we manually mask a portion of the signals by setting the value to zeros  within several masking blocks. 
    \item \textbf{\textit{Sampling rate mismatching.}} Physical biosignals are continuous (analogy), which need to be discretised and stored as digital signals by a given sampling rate for further utilisation. A very low sampling rate can lead to information loss~\cite{jerri1977shannon}. To simulate this, we randomly down-sample some frames in the raw biosignals for testing. 
\end{itemize}

\section{Evaluation}
\subsection{Experiment Design}
Since there is no ground truth for uncertainty, it is not straightforward to evaluate the quality of the uncertainty. Our proposed evaluation protocol is to add controllable perturbations to the original biosignals in the testing sets, and then compare the yielded predictive uncertainty under different shifting degrees.

Overall, we define a shifting degree from 0 to 5  with 0 denoting the original testing set. For mixing Gaussian noise and background noise, shift degrees of 1, 2, 3, 4, and 5 indicate an SNR of 50, 40, 30, 20, 10, respectively. For amplitude distortion, threshold of 80\%, 60\%, 50\%, 20\%, 10\% of the maximum amplitude are applied. For signal segment missing, we mask 20\%, 35\%, 50\%, 65\%, 80\% of the raw signals, and for sampling rate mismatching, every 1/80, 1/50, 1/30, 1/20, 1/10 of the data points are evenly dropped. Illustrative examples can be found in Figure~\ref{fig:examples} in Appendix~\ref{apd:shift}.

\vspace{-5pt}
\subsection{Metrics}
To assess a model's performance, classification accuracy$\uparrow$ (arrows indicating which direction is better) is often used, which only concerns its categorical output $\hat{y}=argmax(p(y|x'))$ as a hard prediction. In addition, in this study, we explore the following metrics, where the predictive uncertainty is taken into account by exploiting the predictive probabilities instead of the hard predictions:
\vspace{2pt}
\\\textbf{Brier Score}$\downarrow$. Brier score measures the distance between the one-hot labels and the predictive probabilities. It is computed as $\frac{1}{|D'|}\sum_i{(1-p(y=y'_i|x'_i,\theta))^2}$.
\\\textbf{ECE}$\downarrow$. Expected calibration error~(ECE) measures the correspondence between the predicted probabilities
and empirical accuracy. It is computed as the weighted average gap between within bucket accuracy and probability. A bucket  $B_s=\{p(y=y'_i|x'_i,\theta) \in(\rho_s,\rho_{s+1})\}$ with $\rho_s$ denotes the quantiles of the predictive probabilities. Hence, $ECE= \sum_{s=1}^S\frac{|B_s|}{|D'|}|acc(B_s)-conf(B_s)|$ with $acc(B_s)$ and $conf(B_s)$ denoting the accuracy and the average predictive probability within $Bs$.
\vspace{2pt}
\\\textbf{Predictive entropy}. For each testing sample, entropy describes the average level of information in a random variable. In this context, this variable is the predictive probability and hence $H(p|\theta)=-\sum_{k=1}^Kp(y=k|x'_i,\theta) log(p(y=k|x'_i,\theta))$, can capture the data uncertainty. For methods including MCDropout, Bayesian, Ensemble, a testing sample will be passed into the model $M$ times, and  predictive entropy is formulated as  $H(p)=-\sum_{k=1}^K(\frac{1}{M}\sum p(y=k|x'_i,\theta_m) log(\frac{1}{M}\sum p(y=k|x'_i,\theta_m))$. This captures both aleatoric and epistemic uncertainty. For an easy illustration, we will the notation \textbf{Uncertainty}, which measures the average predictive entropy across the whole testing set $D'$. 
\vspace{3pt}
\\\textbf{What do we expect to see?}
Intuitively, on increasingly shifted data,  a model's performance might degrade, reflected by a decrease in \acc and a rise in \textit{Brier}. Moreover, ideally, this decrease of performance should coincide with an increase in \textit{Uncertainty}. In particular, an increased uncertainty implies that the model becomes less and less confident of its predictions, and this will be a good indicator of potential dataset shifts during inference for real-world health applications. Meanwhile, we would expect a good model remains well-calibrated under different dataset shifts, which is represented by a small and stable ECE.
\section{Results and Analysis}
In this section, we present and discuss the uncertainty estimation results achieved on the three health tasks separately, and then summarise our key findings associated with the uncertainty under dataset shift for biosignals.

 
 \begin{figure*}[t]
  \centering
  {
 \subfigure[Gaussian shift.]{%
    \centering
     \label{fig:COVID1}
     \includegraphics[width=0.275\textwidth]{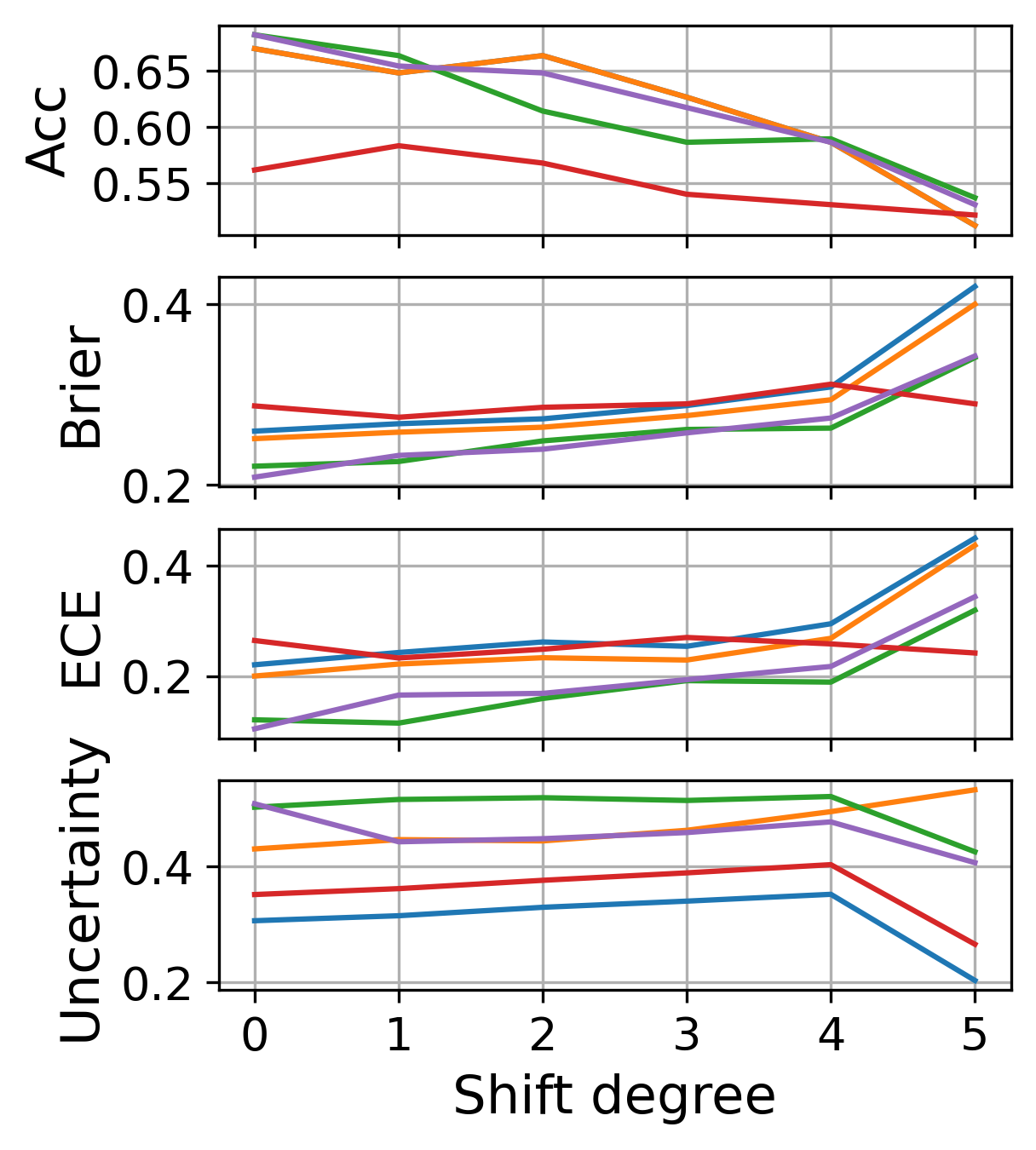}}
 \subfigure[Background noise.]{%
    \centering
     \label{fig:COVID2}
     \includegraphics[width=0.275\textwidth]{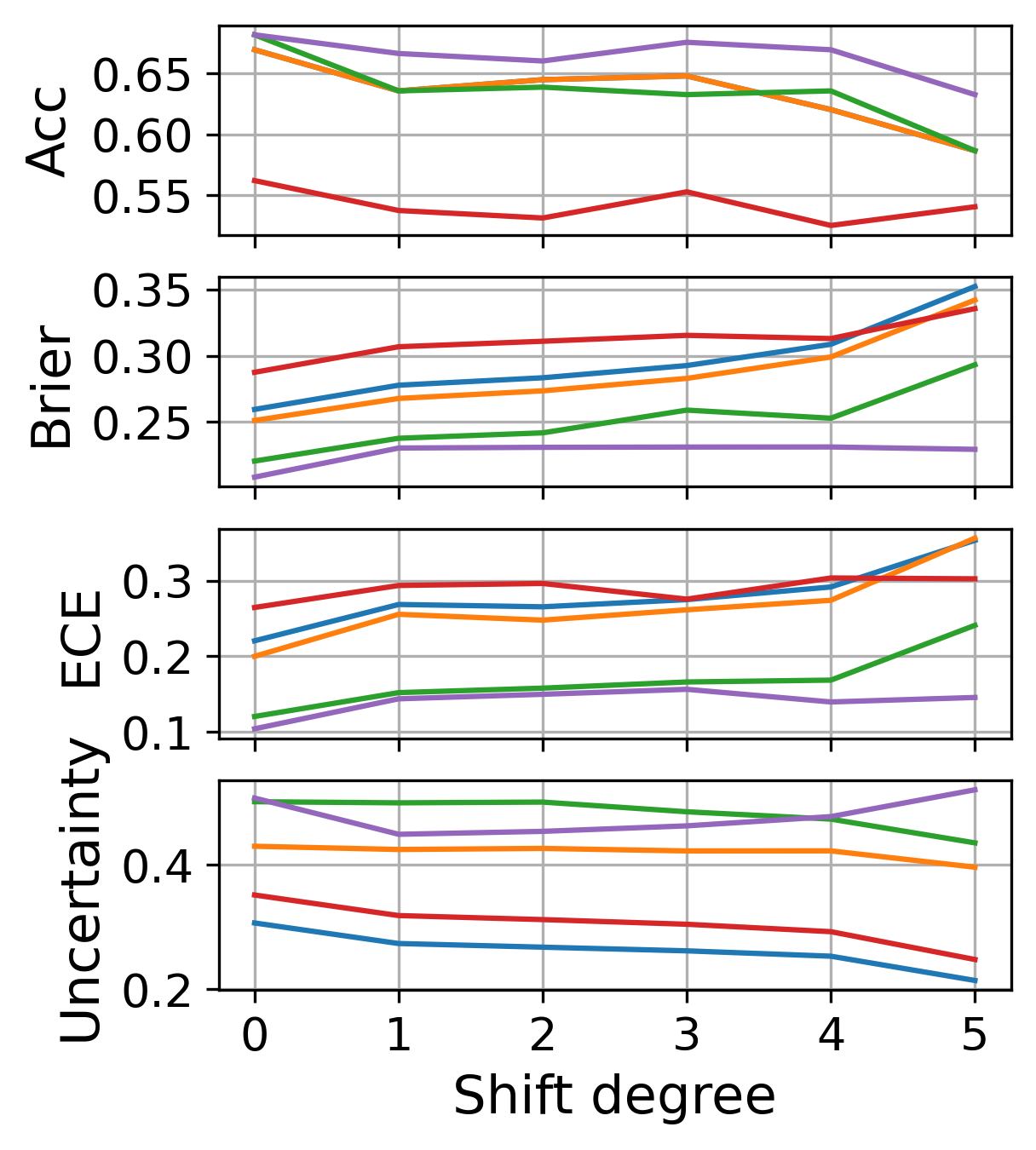}}
 \subfigure[Distortion shift.]{%
    \centering
     \label{fig:COVID3}
     \includegraphics[width=0.395\textwidth]{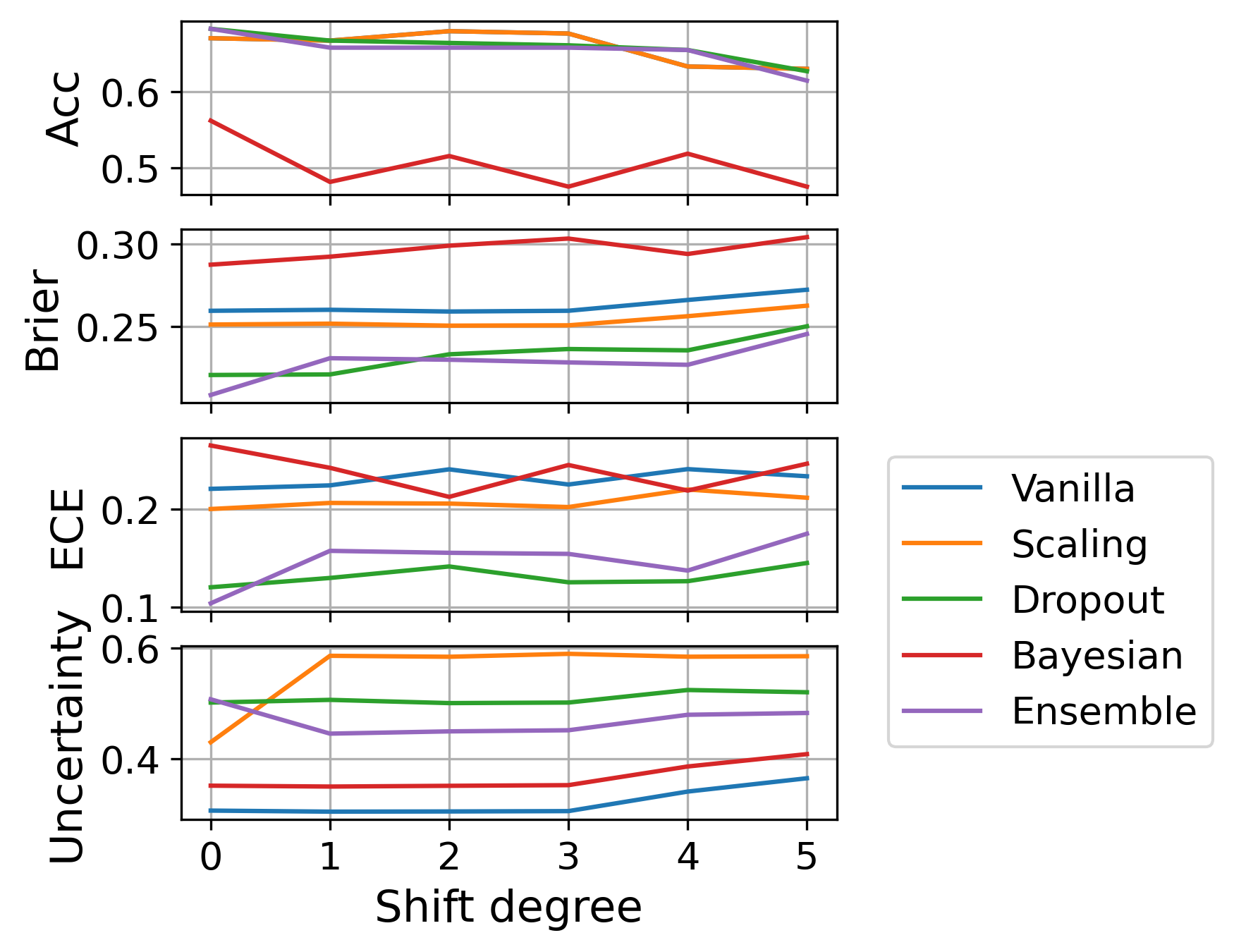}}
     }   
     \vspace{-4pt}
 \caption{Accuracy and uncertainty under various corruptions for COVID-19 detection task. Note that Vanilla and Scaling methods yield the same Acc, so their lines overlapped (and only the Scaling one shows up)} \label{fig:COVID} 
\end{figure*}

\begin{figure}[t]
  \centering
  {
 \subfigure[Gaussian shift.]{%
    \centering
     \label{fig:COVID_b1}
     \includegraphics[width=0.22\textwidth]{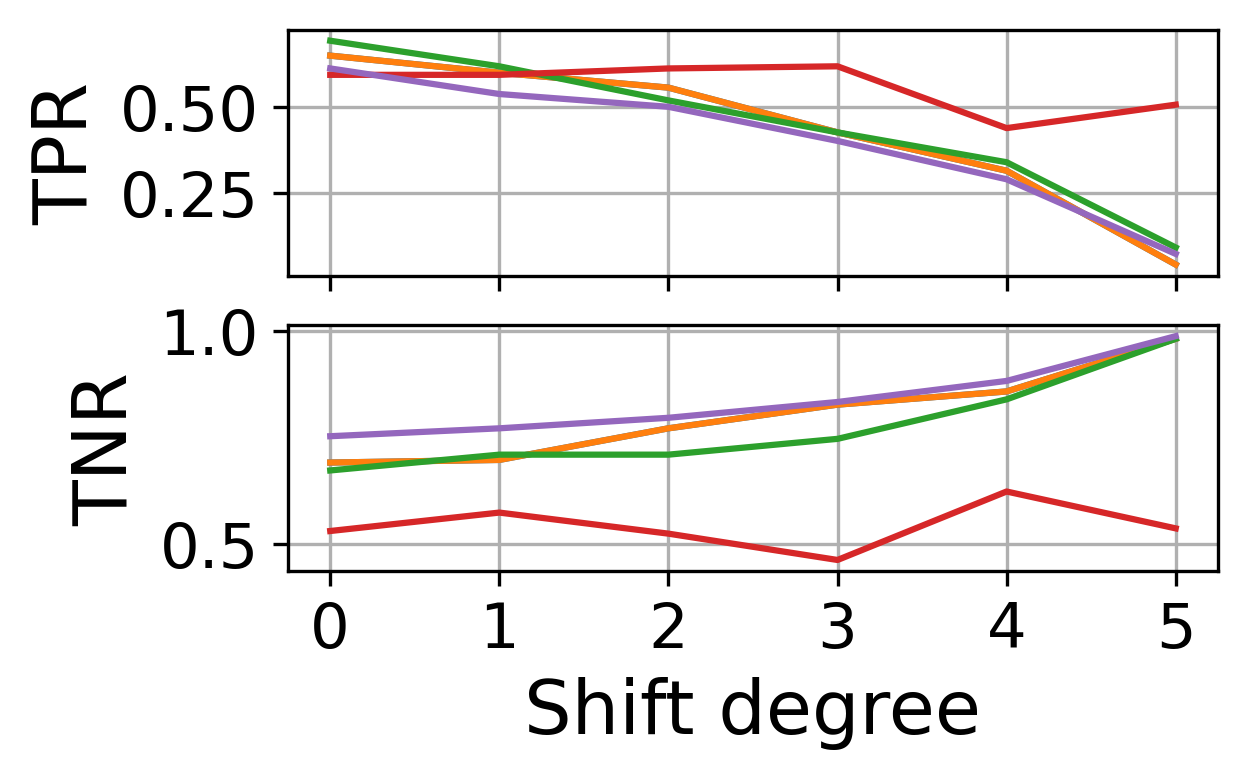}}
 \subfigure[Background noise.]{%
    \centering
     \label{fig:COVID_b2}
     \includegraphics[width=0.22\textwidth]{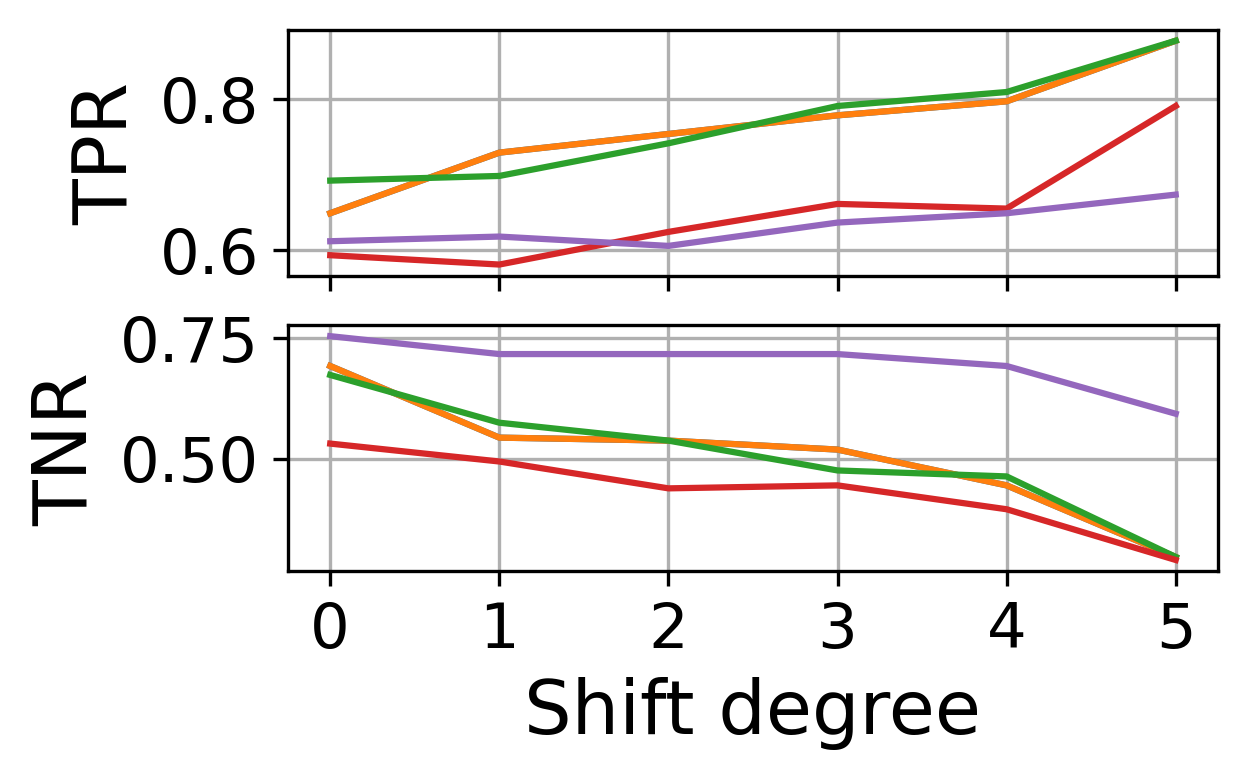}}
  }
  \vspace{-4pt}
 \caption{True positive rate (TPR) and true negative rate (TNR) on the increasing shift for COVID-19 task, with the same colour legend used in Figure~\ref{fig:COVID}.} \label{fig:COVID_b}  
 \vspace{-15pt}
\end{figure}

\subsection{Results for COVID-19 prediction task}
Experimental results for COVID-19 prediction are shown in Figure~\ref{fig:COVID}.
First, from Figure~\ref{fig:COVID1},~\ref{fig:COVID2}, and~\ref{fig:COVID3}, it is clear that all methods achieve worse performance with the increased shift degree, as \acc decays significantly with \Brier and \ECE showing an upward trend. 
Yet, \Uncertainty does not perform as expected for all the cases: in Figure~\ref{fig:COVID1} on degree 5 and in Figure~\ref{fig:COVID2} for all degrees, most methods yield declining uncertainties. While an increasing \Brier implies that a model becomes more and more uncertain, the corresponding \Uncertainty decreases, indicating that the model produces over-confident incorrect predictions. 

Moreover, we observe that the deterministic model might lead to biased predictions under severe dataset shift. In Figure~\ref{fig:COVID_b}, we inspect the true positive rate~(TPR) and true negative rate~(TNR) in this binary classification task. Figure~\ref{fig:COVID_b1} shows that with severer Gaussian noise, all methods except Bayesian tend to classify more testing samples into the negative group, while the opposite direction can be observed in Figure~\ref{fig:COVID_b2} with TV show noise. It is worth highlighting that a balanced testing set is used in this task, otherwise, \acc may not be a good metric to evaluate a model's generalisation performance. For example, if the negative class is the majority in the testing set, a severer Gaussian noise shift would result in higher \textit{Accuracy}. This is consistent with the finding that uncertainty-unaware dataset shift evaluation can be misleading, as suggested by~\cite{band2021benchmarking}.

Comparing the five methods, Ensemble is relatively the best regarding both \acc and \textit{Uncertainty}. The post-hoc calibration method (Scaling) cannot keep \textit{ECE}, as the temperature scale factor $T$ was optimised on non-shifted data and thus the model was are probably not tolerant of various types and degrees of dataset shift. In contrast, although Bayesian achieves the lowest \textit{Accuracy}, its \ECE shows as only a small fluctuation. This might due to the fact that the size of training data for this task is very small but the parameter estimation of the Bayesian model is more computationally intractable and needs more training data, and thus the model is still under-fitted.

\subsection{Results for respiratory task} 
\begin{figure*}[t]
  \centering
  {
 \subfigure[Gaussian shift.]{%
    \centering
     \label{fig:Res1}
     \includegraphics[width=0.28\textwidth]{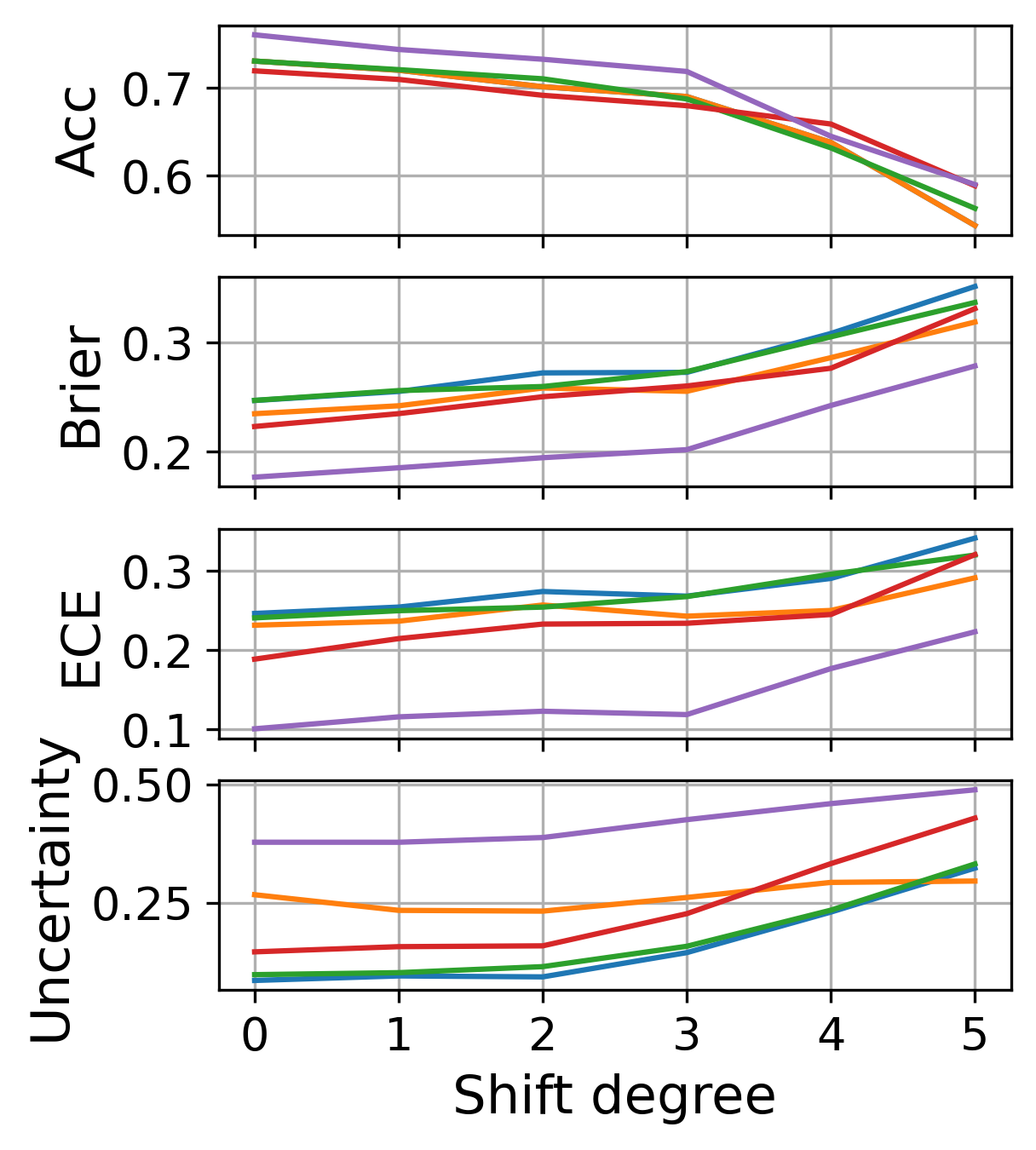}}
 \subfigure[Distortion shift.]{%
    \centering
     \label{fig:Res2}
     \includegraphics[width=0.272\textwidth]{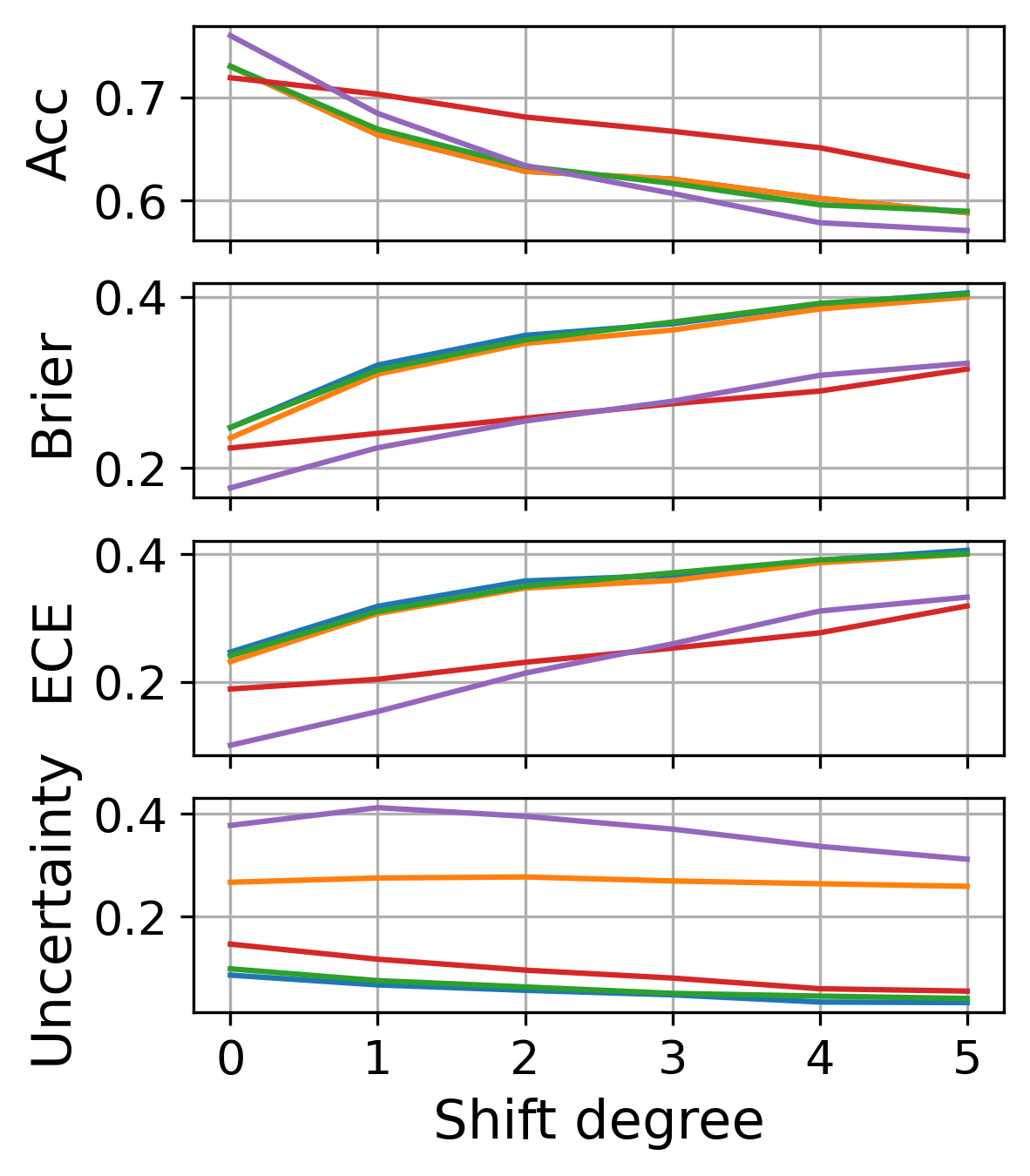}}
 \subfigure[Missing shift.]{%
    \centering
     \label{fig:Res3}
     \includegraphics[width=0.395\textwidth]{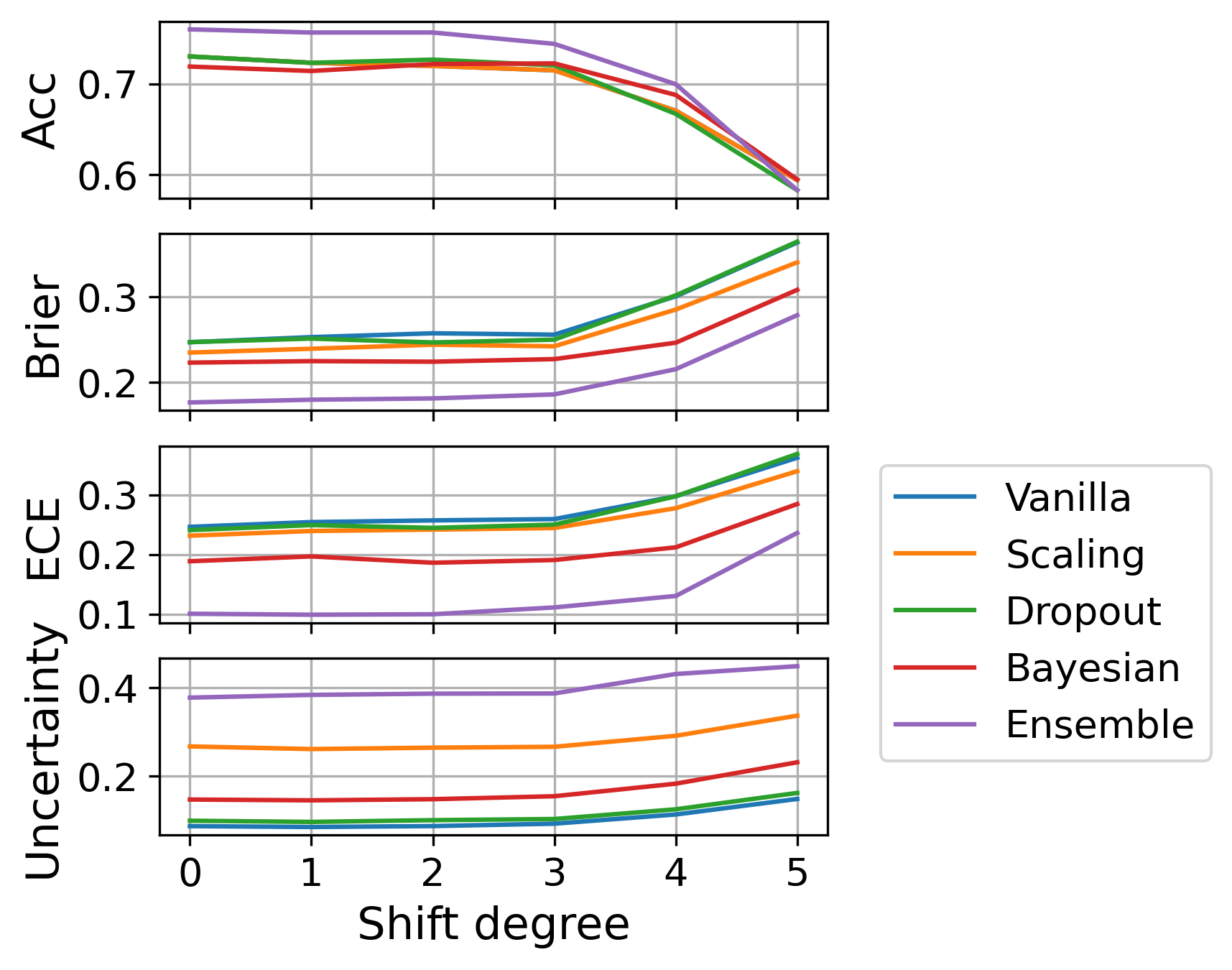}}
     }   
     \vspace{-4pt}
\caption{Accuracy and uncertainty under various corruptions for respiratory abnormality detection task.}\label{fig:Res}  
\end{figure*}

\begin{figure*}[t]
  \centering
  {
 \subfigure[Selective prediction.]{%
    \centering
     \label{fig:RESa}
     \includegraphics[width=0.29\textwidth]{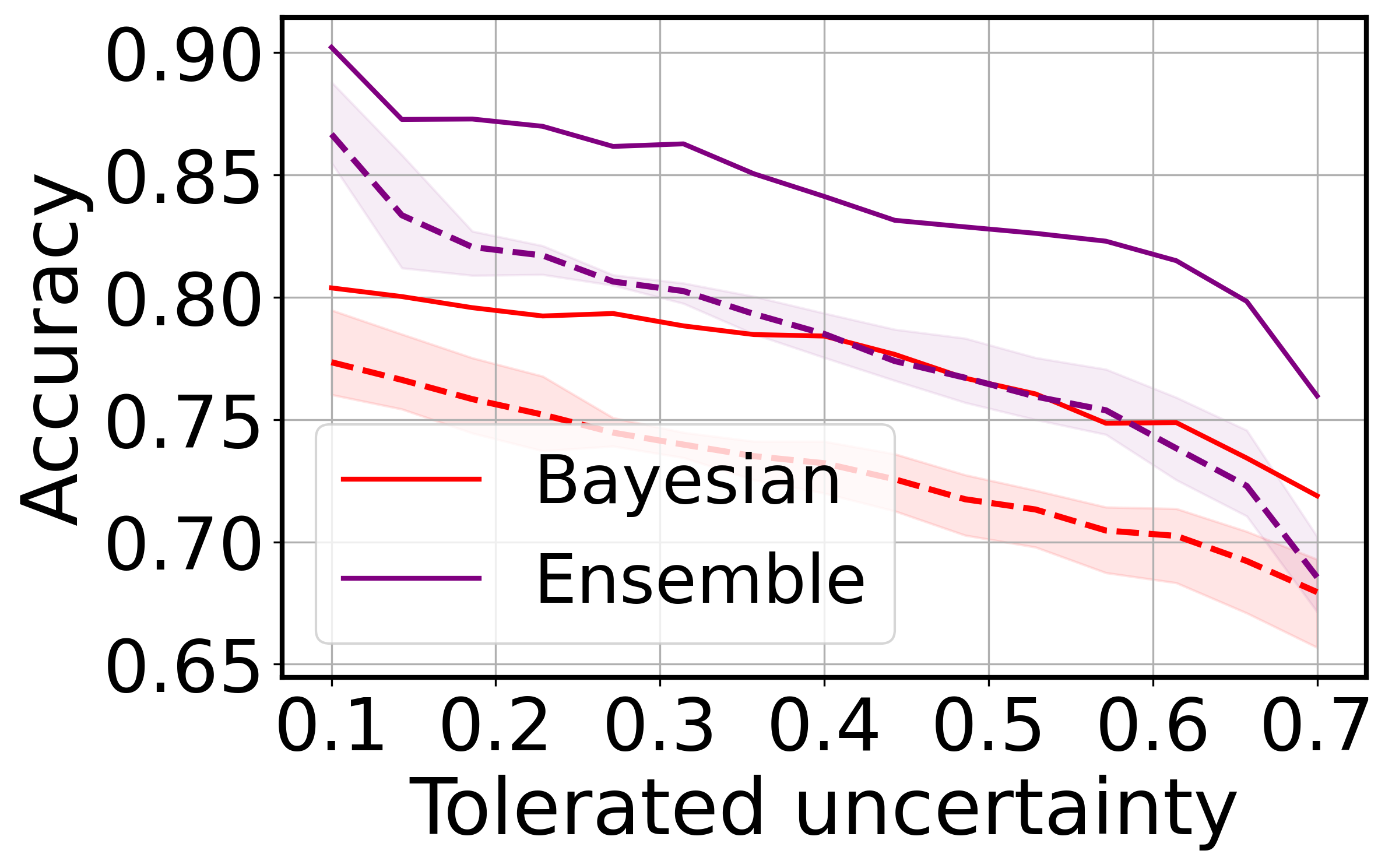}}
 \subfigure[Shift detection.]{%
    \centering
     \label{fig:RESb}
     \includegraphics[width=0.29\textwidth]{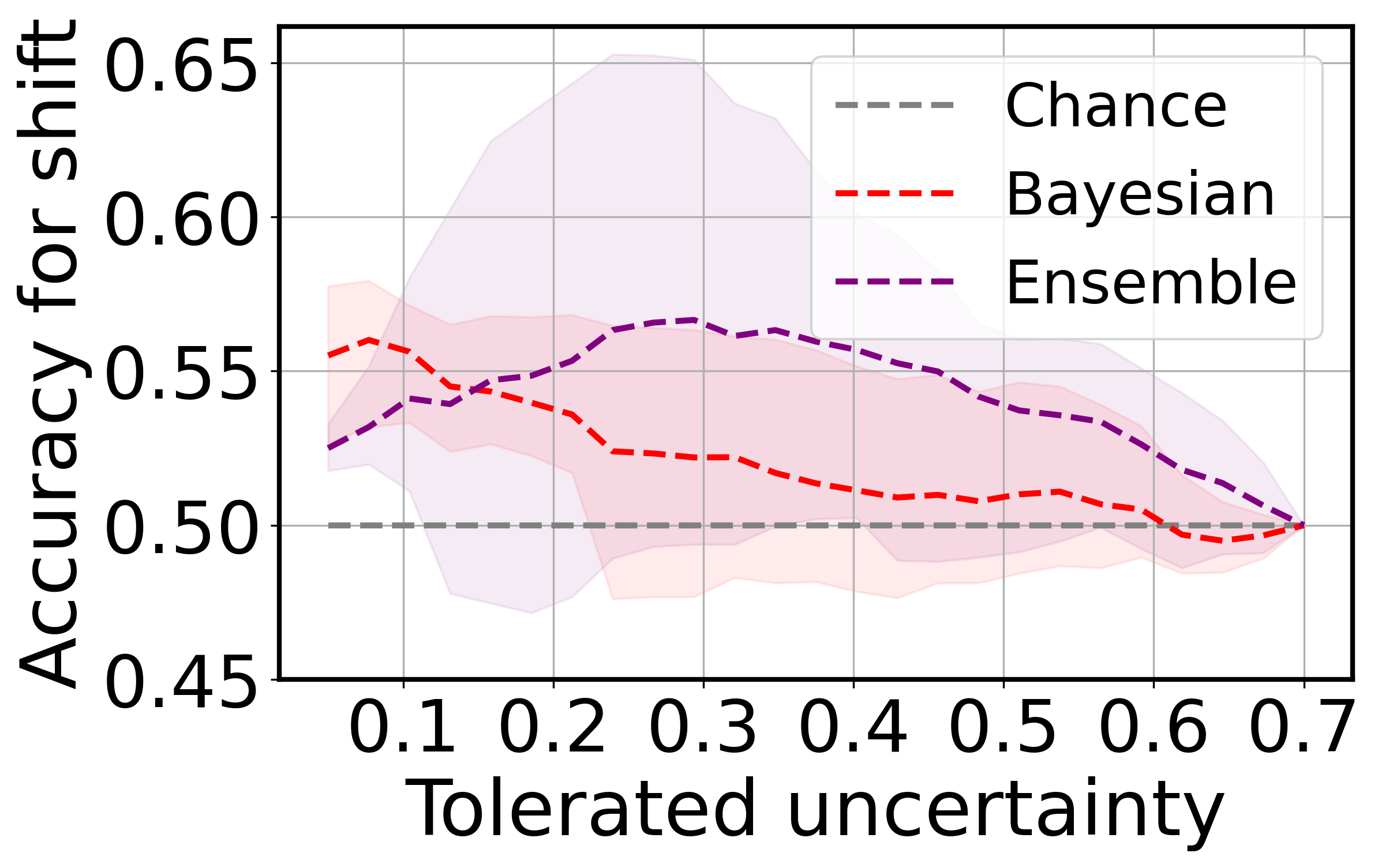}}
  \subfigure[A case: uncertainty from Ensemble method.]{%
    \centering
     \label{fig:RESc}
     \includegraphics[width=0.28\textwidth]{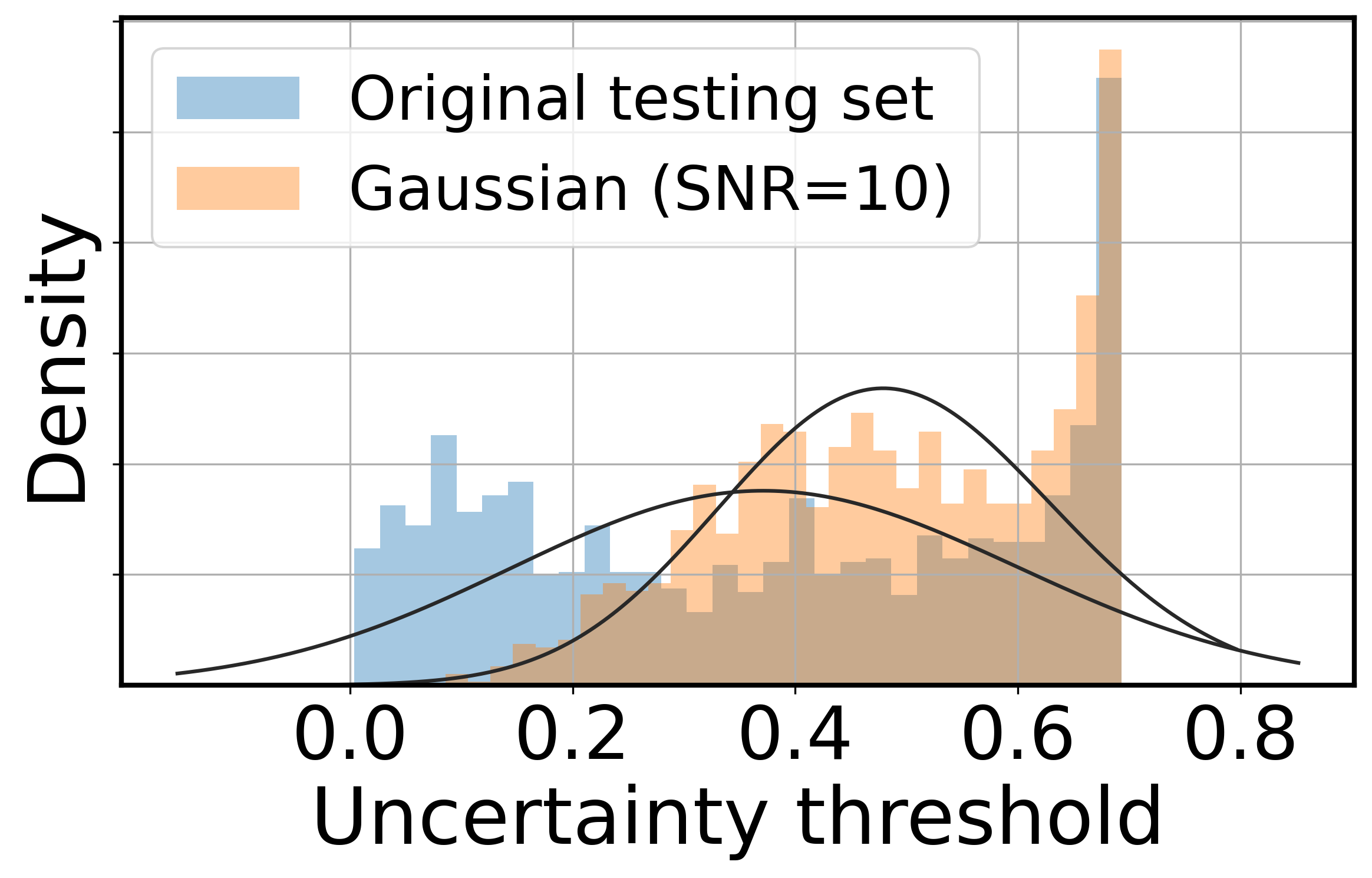}}
     }
     \vspace{-10pt}
 \caption{A detailed comparison for the quality of uncertainty in the respiratory abnormality detection task. (a) \acc on the remained data with samples having uncertainty higher the threshold referred. Solid lines denote the original testing set, while dashed lines present the average accuracy on the mixed original and shifted sets with shade showing the variance among shift degrees and types. (b) Accuracy for shift detection on the mixed original and shifted testing set:  samples with $Predictive Entropy>threshold$ will be detected as shifted inputs. (c) \Uncertainty distribution from Ensemble method on Gaussian shift with $degree=5$.} \label{fig:RES2}  
 \vspace{-15pt}
\end{figure*}

In this task, as the dataset shift gets severer, for all methods, \acc declines, \Brier becomes larger, and for most cases, \Uncertainty goes up, indicating that the models are getting more uncertain, as what we expected. It is also worth noting that although \acc of those methods are relatively close, \Brier gives a clear and fine-grained picture showing that the output probabilities are quite different and Ensemble can achieve the minimum error.
However, an increase in \ECE can be observed from Figure~\ref{fig:Res1},~\ref{fig:Res2}, and~\ref{fig:Res3}, which shows that all methods yield over-confident predictions on the increasingly shifted testing set. 

Since \Brier and \Uncertainty show an upward trend in this task, we conduct further in-depth analysis to inspect whether the quantified predictive uncertainty, particularly by Bayesian and Ensemble approaches, are sufficient to secure the predictions under dataset shift. First, as uncertainty is usually used to select low-confident predictions and pass them to doctors~\citep{leibig2017leveraging}, we compare the \acc on the remained data in Figure~\ref{fig:RESa}. Despite the performance improvement from selective prediction, the gap between the \acc on the original testing set (solid lines) and on the data mixed with shifts (dashed lines) is notable. This implies that the estimated distributional uncertainty by Bayesian and Ensemble models cannot help to hold the impressive accuracy achieved on the non-shift testing set. Figure~\ref{fig:RESb} further verifies the incapability, where we investigate if \Uncertainty can be exploited to detect a shifted input from the training distribution. Yet, the performance is just slightly better than random guess: $accuracy<0.6$ of Bayesian and Ensemble \textit{v.s.} 0.5 of chance level. This indicates that the shifted data is not distinguishable from the original data by the present uncertainty estimation methods.
Further, in Figure~\ref{fig:RESc}, we display an example of the severest Gaussian shift with blue bars denoting the uncertainty distribution on the original testing set and orange bars for the shifted set. It is good to see that the predictive uncertainties on the shifted set are generally higher than that on the non-shifted set. However, those two distributions are still close to each other, which undermines the capacity of current uncertainty measure approaches to tackle dataset shift in real applications.

\subsection{Results for heart arrhythmia task}
\begin{figure*}[t]
  \centering
  {
 \subfigure[Gaussian shift.]{%
    \centering
     \label{fig:ECG1}
     \includegraphics[width=0.285\textwidth]{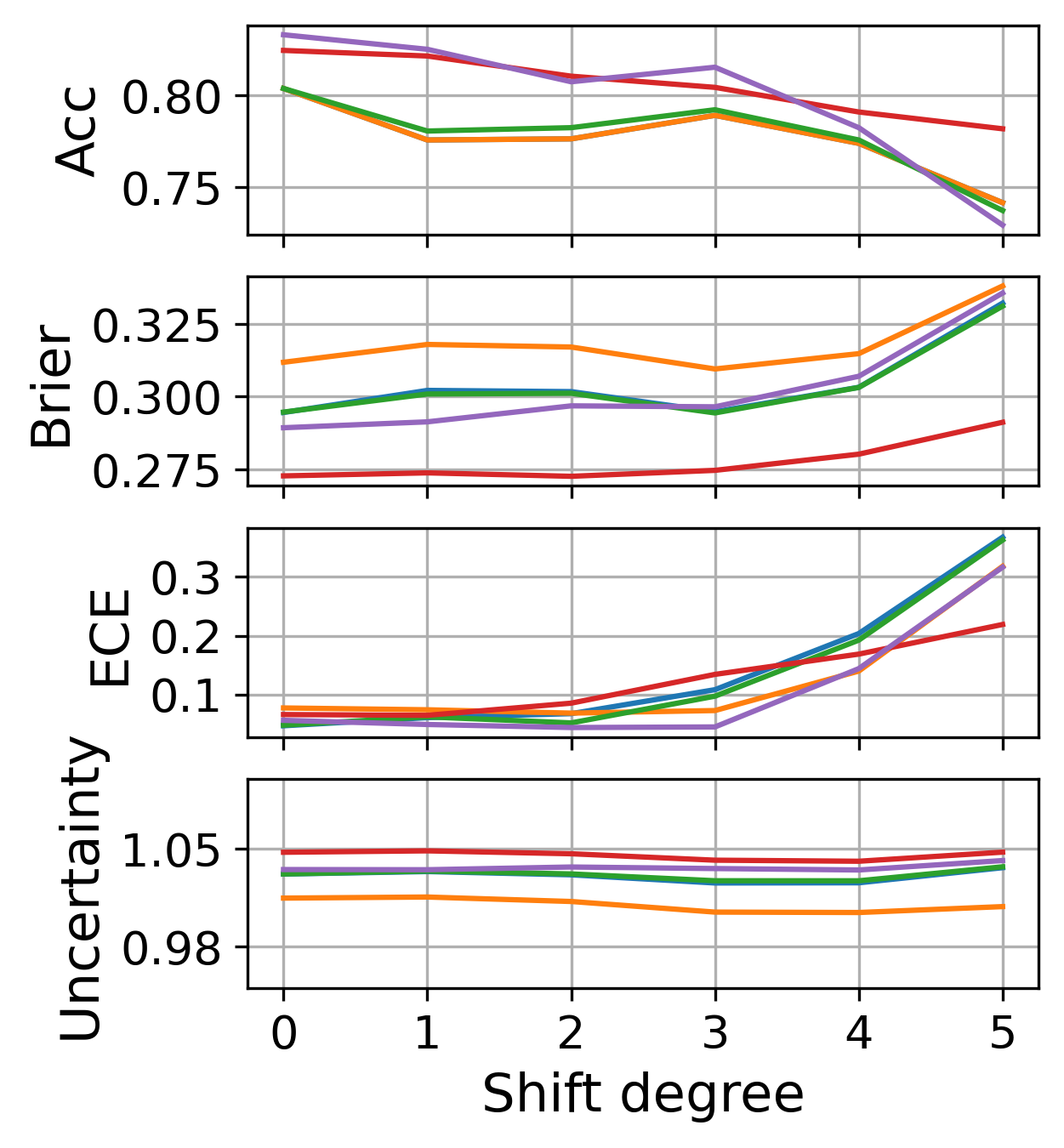}}
 \subfigure[Missing shift.]{%
    \centering
     \label{fig:ECG2}
     \includegraphics[width=0.275\textwidth]{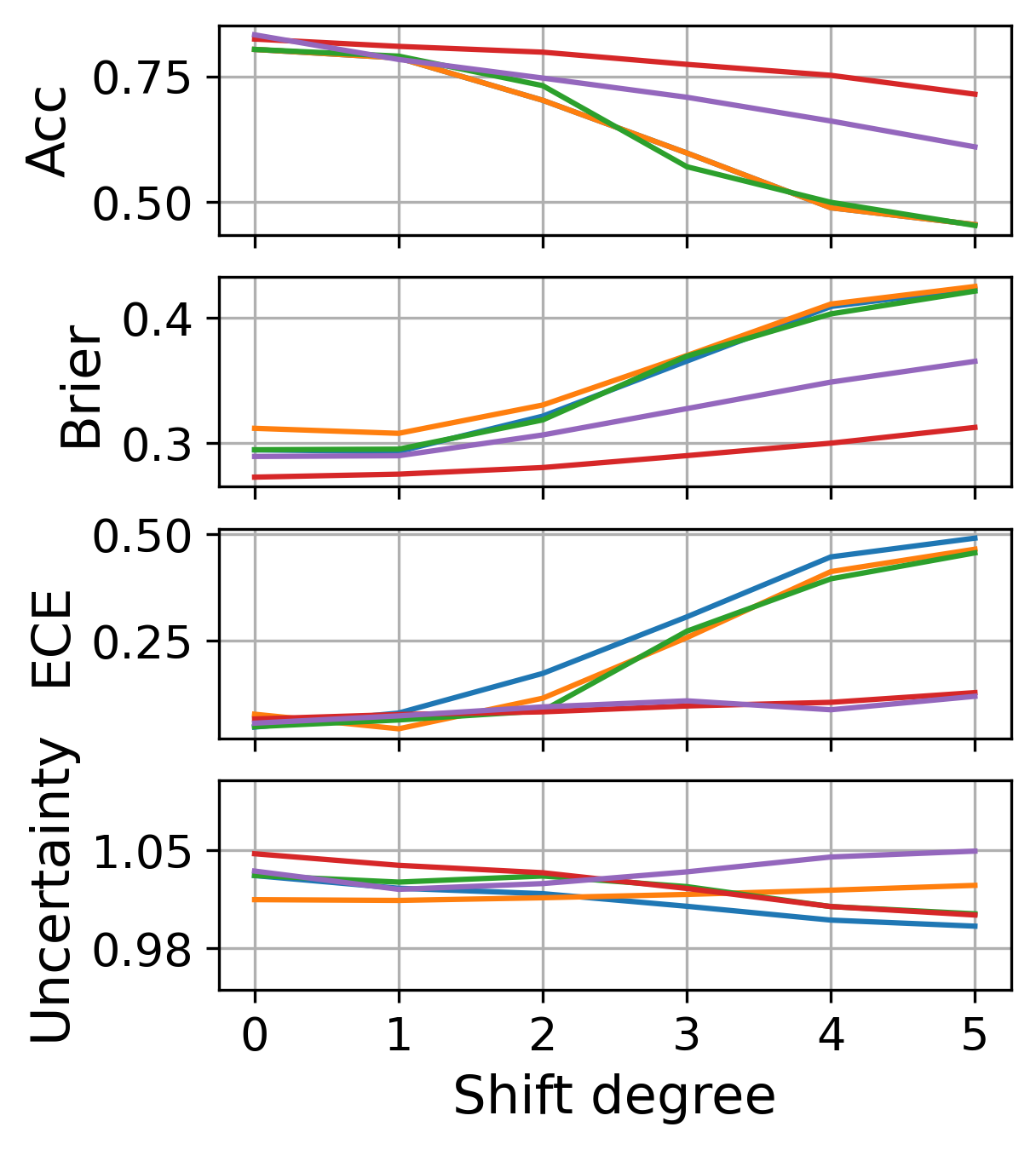}}
 \subfigure[Sampling rate mismatch.]{%
    \centering
     \label{fig:ECG3}
     \includegraphics[width=0.405\textwidth]{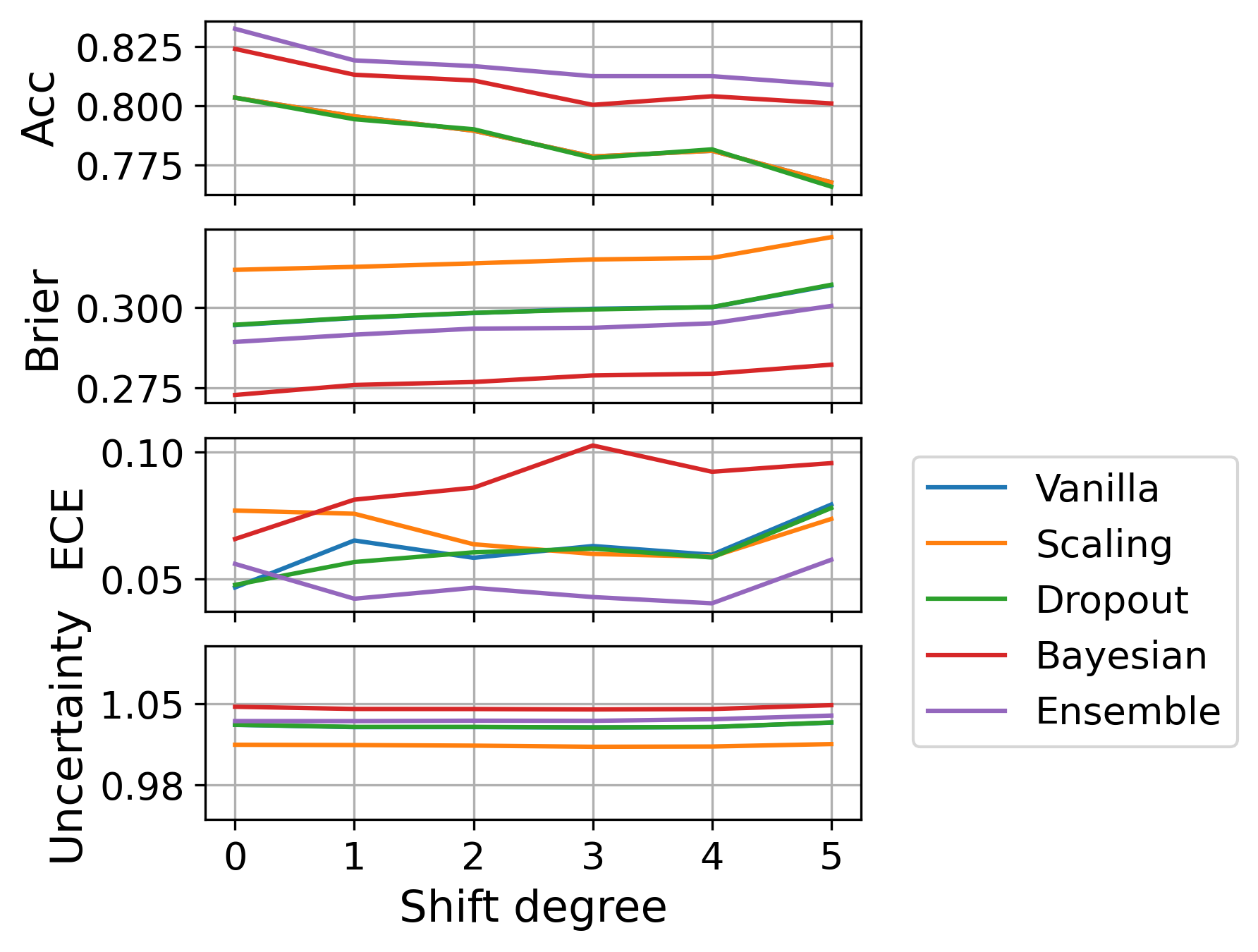}}
     }
     \vspace{-10pt}
 \caption{Accuracy and uncertainty under various corruptions for heart arrhythmia detection.} \label{fig:ECG}  
\end{figure*}

\begin{figure*}[t]
  \centering
  {
 \subfigure[A correct prediction example.]{%
     \label{fig:ECG_ex1}
     \includegraphics[width=0.48\textwidth]{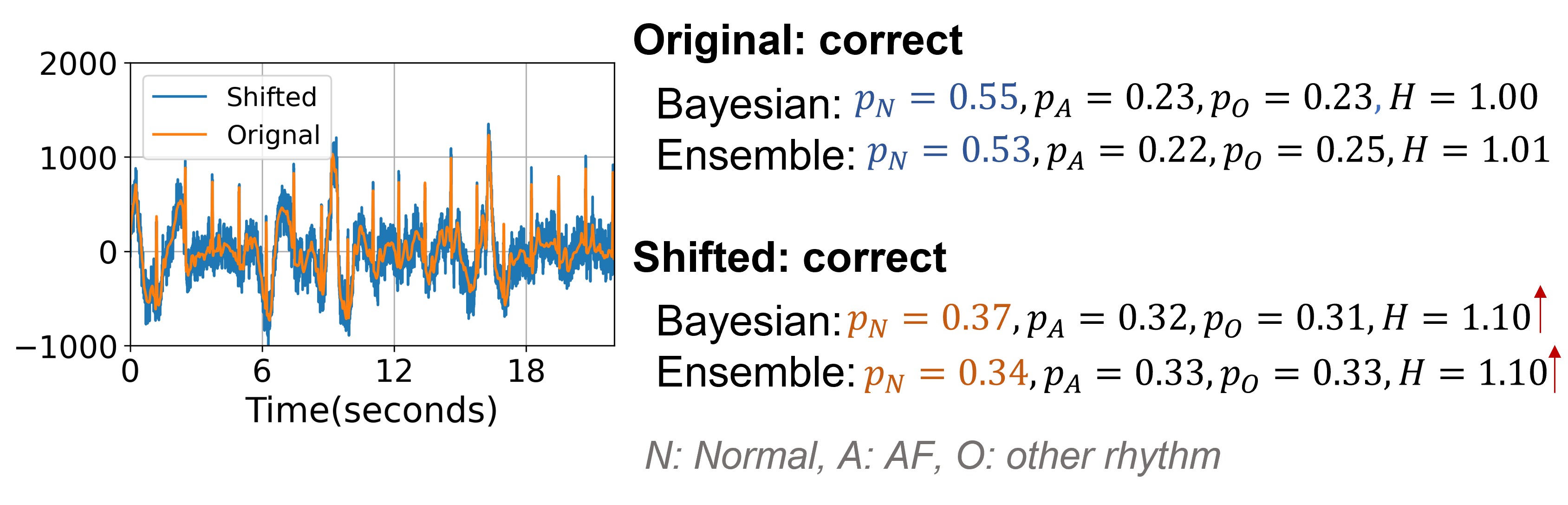}}
 \subfigure[An incorrect prediction example.]{%
     \label{fig:ECG_ex2}
     \includegraphics[width=0.48\textwidth]{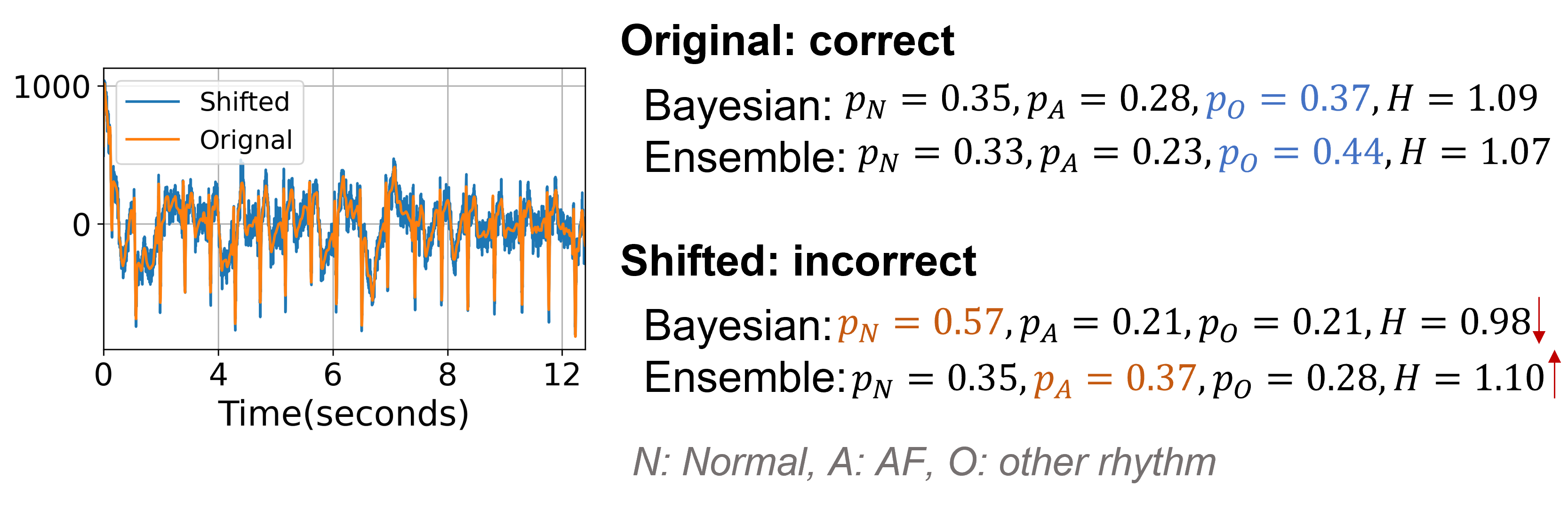}}
     }
 \caption{Comparison of output probability and predictive uncertainty from two methods on original and shifted samples. Gaussian noise is added to the signals with $SNR=10$. (a) shows a normal ECG recording, and (b) is a non-AF abnormal rhythm recording. } \label{fig:ECG_ex} 
 \vspace{-10pt}
\end{figure*}

Results for the heart arrhythmia detection task on Gaussian noise shift, segment missing shift, and sampling rate mismatch shift are presented in Figure~\ref{fig:ECG}. The primary findings are consistent with the observations from the prior two tasks: \acc degrades and model becomes increasingly over-confident for all methods. Although as shown in Figure~\ref{fig:ECG2}, with a great proportion of the signals missed, Bayesian and Ensemble methods can keep relatively good \acc compared to other methods, the slight increase in \ECE and the small reduction in \Uncertainty suggest that the uncertainty might be not fully reliable.

We also carry out some visualisation comparison in Figure~\ref{fig:ECG_ex}. 
Figure~\ref{fig:ECG_ex1} presents an example where Bayesian and Ensemble both achieve correct predictions for the original and shifted inputs. This normal rhythm ECG signal is predicted as normal with a probability over 0.5, but with the Gaussian noise shift, Bayesian and Ensemble yield lower probability for the normal class: $p_N=0.37$ and $p_N=0.34$, respectively. Meanwhile, \Uncertainty of those two approaches rises from $H=1.0$ to $H=1.1$. Herein, this is a case that the distributional shift has been captured by the predictive uncertainty. In contrast, Figure~\ref{fig:ECG_ex2} demonstrates a failed prediction on the shifted input, as the shifted non-AF abnormal rhythm sample is predicted as a normal and an AF signal by Bayesian and Ensemble, respectively. What's worse, Bayesian model yields a lower uncertainty value: from $H=1.09$ to $H=0.98$, which indicates that it is less uncertain although it makes a wrong prediction.
This case may lead to a negative impact in a real automatic diagnosis system, as the estimated uncertainty falls short of its promise to reflect the reliability of the prediction. 

\subsection{Summary and Takeaways}
Combining the observations from all three tasks, we draw the following conclusions and recommendations,
\begin{itemize}
    \item 
    With increasing dataset shift in biosignals, all uncertainty estimation approaches we evaluated fail to report a reasonable increasing uncertainty score to notify the changes in data distribution, while the performance in terms of accuracy degrades sharply.
    \item Ensemble can achieve a slightly better uncertainty estimation than the other methods, although it needs relatively heavier computing cost and memory consumption. Bayesian method can obtain similar performance when training data is sufficient. 
    \item Classifiers trained on non-shifted data might be biased on a specific dataset shift during inference. Thus, the measure of prediction uncertainty is as important as the prediction itself, particularly in safety-critical healthcare applications.
    \item Models may become more and more over-confident as the shift gets severer. None of the existing methods is perfect in capturing distributional shifts and calibrating the deep neural networks. New approaches are needed.
\end{itemize}

\vspace{-10pt}
\section{Conclusions}
\label{sec:con}
In this paper, we conduct extensive experiments and analysis to assess the estimated predictive uncertainty under dataset shift on biosignals. We implemented five uncertainty quantification methods on three representative biosignal classification tasks under a controlled dataset shift. To enable a comparison, we propose a protocol to analyse all methods without requiring new data: we synthesise signal-specific distributional shifts according to real signal data collection scenarios. Our work establishes a benchmark for future evaluation of uncertainty quantification methods.


\bibliography{ref}

\appendix
\clearpage
\section{Appendix}
\subsection{Task implementation}\label{apd:task}
We introduce the details of our model implementation as below,
\vspace{2pt}
\\\textbf{COVID-19 prediction.} This is a binary classification task, where cough, breathing, and voice sounds are transferred into spectrograms to distinguish COVID-19 positive from negative participants. VGGish~\citep{hershey2017cnn} is leveraged as the backbone model. VGGish is a convolutional neural network, followed by a two dense layer-based classifier. We insert the dropout layer after each dense layer with a drop rate of 0.5. This model is implemented by Tensorflow 1.18, and thus for Bayesian method, lib \textit{Tensorflow Probability}\footnote{\url{https://www.tensorflow.org/probability/api_docs/python/tfp/layers/DenseFlipout}} is applied.
\vspace{2pt}
\\\textbf{Respiratory abnormality detection.} This binary classification task is to detect whether a breathing sound segment contains abnormalities, including crackle and wheeze. We use the backbone of a deep convolutional model (ResNet) proposed in~\cite{gairola2020respirenet}. Dropout ($p=0.5$) and batch normalisation are leveraged to reduce over-fitting. This task is implemented by PyTorch 2.0 with \textit{BLiTZ} (a Bayesian neural network library)\footnote{\url{https://towardsdatascience.com/blitz-a-bayesian-neural-network-library-for-pytorch-82f9998916c7}}.
\vspace{2pt}
\\\textbf{Heart arrhythmia detection.} The task is to predict, from a single short ECG lead recording (between 30\,s and 60\,s in length), whether the recording shows normal sinus rhythm, atrial fibrillation (AF), or an alternative rhythm. The backbone is a transformer-based three-class classification model with drop rate $p=0.15$. This task is also implanted by Tensorflow 1.18 with \textit{Tensorflow Probability}.

All datasets are publicly available\footnote{The COVID-19 dataset is available upon request with a data transfer agreement.} and model implementation is based on the released code for fairness and reproducibility. For all tasks, we split the entire dataset into training/validation/testing sets by the ratio of 7:1:2. For each shifting type/degree, we create a new testing set of the same size by adding a specific perturbation to testing samples one by one. For temperature scaling, we adjust parameter $T$ on the validation set for three tasks, and finally use $T=1.5$, 1.2, and 1.15, respectively. We train $M=10$ models for ensemble and similarly, pass the input to the Bayesian and MCDropout models for $M=10$ times. However, we notice that  $M>5$ is good enough to achieve very close performance. All codes will be publicly available on Github.
\newpage
\subsection{Synthetic shifts}\label{apd:shift}
Signal to noise ratio (SNR) can be defined as follows,
\begin{equation} \label{equ:SNR}
 SNR = 10log\frac{RMS^2_{signal}}{RMS^2_{noise}},
\end{equation}
where $RMS$ denotes the root mean square value, that is $RMS=\sum{s_i^2}$.

For Gaussian noise shift, we generate a signal with the same length $l$ to original signal following Normal distribution as,
\begin{equation}
Noise_{Gaussian}, n_i \sim \mathcal{N}(0,\delta), 
\end{equation}
and thus 
\begin{equation}
 \delta = \sqrt{\frac{n_i^2}{l}}, 
\end{equation}
 
For a clip of background signal $Noise_{Back}$, a coefficient $\lambda$ is used to scale the recording, 
\begin{equation}
SNR = 10log\frac{\sum{s_i^2}}{\sum{(n_i*\lambda)^2}}.
\end{equation}
Herein,
\begin{equation}
 \lambda = \frac{\sum{s_i^2}/SNR}{\sum{n_i^2}}.
\end{equation}

Finally, the shifted signal is formulated as,
\begin{equation}
Signal_{Shifted} = Signal  +  Noise.
\end{equation}

\begin{figure}[t]
  \centering
  {
 \subfigure[COVID task: shift on cough recordings.]{%
    \centering
     \label{fig:a}
     \includegraphics[width=0.5\textwidth]{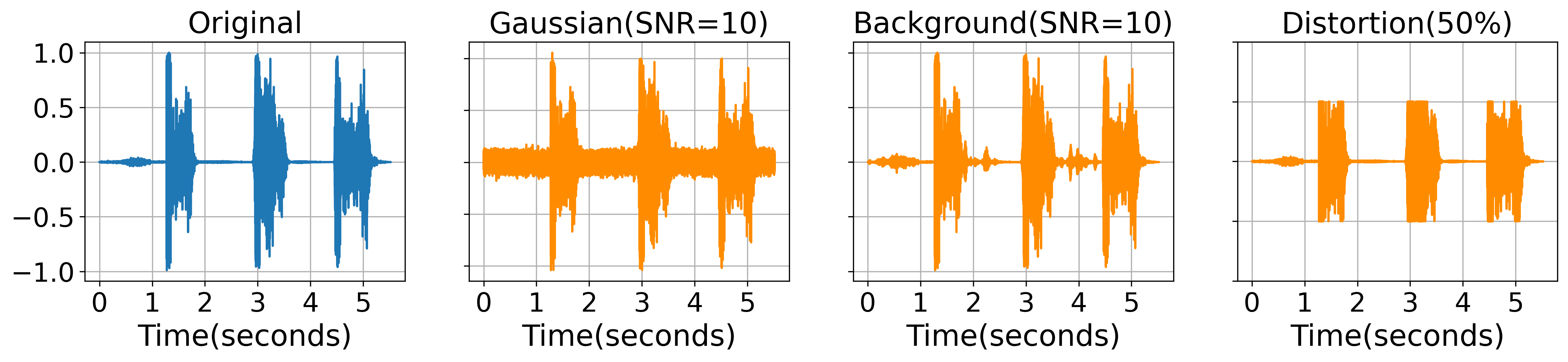}}
 \subfigure[Respiratory task: shift on breathing recordings.]{%
    \centering
     \label{fig:b}
     \includegraphics[width=0.5\textwidth]{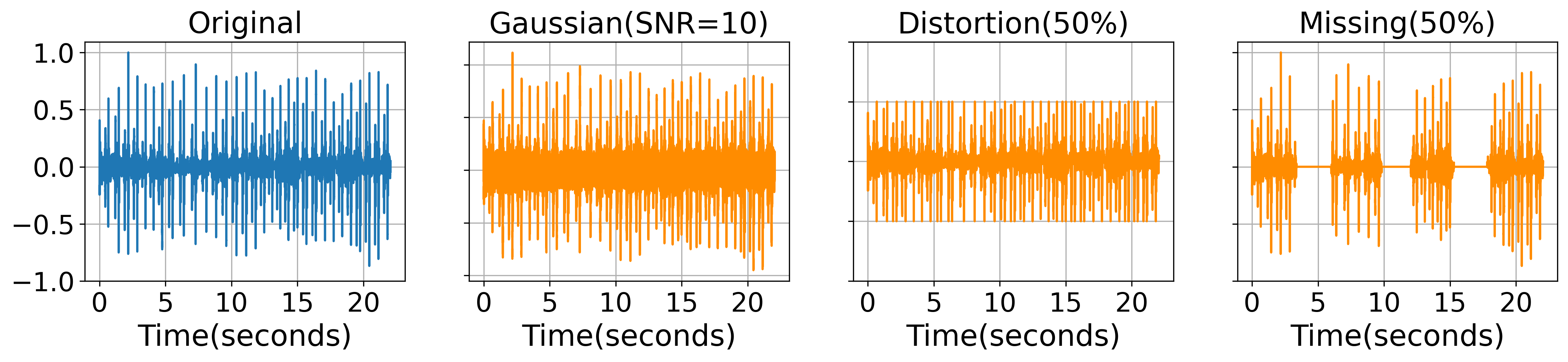}}
  \subfigure[Heart task: shift on ECG recordings.]{%
    \centering
     \label{fig:c}
     \includegraphics[width=0.5\textwidth]{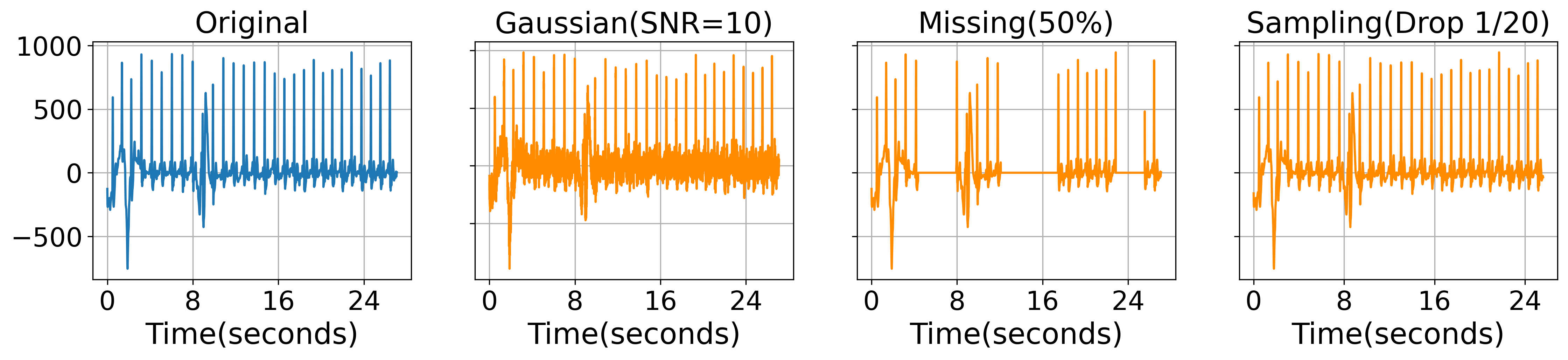}}
     }   
 \caption{Illustrations of synthesising different specific dataset shifts on biosignals. Blue signals present random selected original signals for the three tasks respectively, and accordingly yellow signals are the synthesised signals under different controlled perturbations with a given degree.} \label{fig:examples}  
\end{figure}

Examples for various types and degrees of dataset shift are given in Figure~\ref{fig:examples}.  

\end{document}